\documentclass[10pt]{article}
\usepackage{graphicx}
\usepackage{amsmath}
\usepackage{amssymb}
\usepackage{amsthm}
\usepackage{epsfig}
\usepackage{mathrsfs}
\usepackage{multirow}
\usepackage[usenames,dvipsnames]{color}
\usepackage{setspace}
\usepackage{algorithm, algorithmic} 
\usepackage{multicol}
\usepackage{subcaption}
\usepackage[margin=2.54cm]{geometry}
\usepackage{lipsum}
\usepackage{bigints}
\usepackage{framed}
\usepackage{colortbl}
\usepackage{calligra}
\usepackage{mathrsfs}
\usepackage{enumerate}
\usepackage{hyperref}
\usepackage{enumitem}
\usepackage[round, sort]{natbib}
\usepackage{cleveref}
\usepackage{xcolor}
\usepackage{mathtools}
\usepackage{dsfont}
\usepackage{adjustbox}
\usepackage{booktabs}

\usepackage{appendix}

\DeclareMathOperator{\diag}{diag}

\DeclareMathAlphabet{\mathcalligra}{T1}{calligra}{m}{n}

\newtheorem{definition}{Definition}

\newtheorem{remark}{Remark}

\newcommand{\bm}{\boldsymbol}

\def\RR{ {\mathbb{R}}}

\newcommand{\mc}[1]{\mathcal{#1}}

\title{
Learning Fair Policies for Infectious Diseases Mitigation using Path Integral Control}
\author{
Zhuangzhuang Jia\thanks{Department of Industrial and Enterprise Systems Engineering, University of Illinois Urbana-Champaign, Urbana, IL 61801, USA. ({\tt zj12@illinois.edu, hyukp2@illinois.edu, gah@illinois.edu}).}
\ \ \ \ \ \ \
Hyuk Park\footnotemark[1]
\ \ \ \ \ \ \
G\"{o}k\c{c}e Dayan{\i}kl{\i}\thanks{Department of Statistics, University of Illinois Urbana-Champaign, Champaign, IL 61820, USA. ({\tt gokced@illinois.edu}).}
\ \ \ \ \ \ \
Grani A. Hanasusanto\footnotemark[1]%
}
\date{}

\begin{document}

\maketitle

\begin{abstract}
Infectious diseases pose major public health challenges to society, highlighting the importance of designing effective policies to reduce economic loss and mortality. 
In this paper, we propose a framework for sequential decision-making under uncertainty to design fairness-aware disease mitigation policies that incorporate various measures of unfairness. Specifically, our approach learns equitable vaccination and lockdown strategies based on a stochastic multi-group SIR model. To address the challenges of solving the resulting sequential decision-making problem, we adopt the path integral control algorithm as an efficient solution scheme. Through a case study, we demonstrate that our approach effectively improves fairness compared to conventional methods and provides valuable insights for policymakers.
\end{abstract}

\section{Introduction}
The COVID-19 pandemic has caused an unprecedented global impact, resulting in over 7 million deaths worldwide~\citep{owid-coronavirus} and causing extensive socioeconomic disruption---including \$3.8 trillion in global consumption losses, the loss of 147 million full-time jobs, and a reduction in income amounting to \$2.1 trillion \citep{lenzen2020global}. Moreover, numerous reports indicate that lower-income communities have been disproportionately affected by the pandemic. \cite{ronkko2022impact} report that the overall income of these communities during the pandemic fell further below pre-pandemic levels compared to the declines experienced by wealthier and average-income communities. In addition, the infection and mortality rates from COVID-19 have been significantly higher in these communities due to socioeconomic disparities, such as higher population density, limited access to private transportation, and reduced availability of timely medical treatments. This growing health inequality presents a critical societal challenge that policymakers must address~\citep{bambra2020covid}.

To address this issue, our work aims to incorporate fairness consideration into the design of disease mitigation policies to ensure the equitable allocation of limited government resources across {\color{black}communities with different socioeconomic statuses}. The proposed approach and our main contributions are summarized as follows:
\begin{enumerate}
    \item We propose a novel framework for sequential decision-making under uncertainty that integrates a stochastic multi-group SIR model with various unfairness penalties, enabling policymakers to design fair vaccination and lockdown strategies. To the best of our knowledge, our work is the first to incorporate the consideration of fairness in the SIR model.
    \item Solving the proposed sequential decision-making problem is computationally challenging with conventional methods, as the inclusion of unfairness penalties and the dynamics of the SIR model results in nonlinear partial differential equations (PDEs). To address the computational challenge, we utilize the path integral control algorithm, an optimization-free scheme that efficiently solves the problem by Monte Carlo sampling. 
    \item We conduct a numerical case study on COVID-19 to derive managerial insights. The results demonstrate that our region-specific policies significantly improve fairness across different socioeconomic regions compared to conventional homogeneous policies (across regions). Notably, our fairness-aware policies suggest prioritizing vaccination efforts in lower-income regions. 

\end{enumerate}

\section{Related Work} 
\paragraph{Modeling Infectious Disease Spread.} 
The classical SIR model, introduced in~\cite{kermack1927contribution}, is a widely used dynamical system for modeling disease evolution, assuming deterministic and homogeneous dynamics across the entire population. To account for uncertainty in disease spread, various stochastic extensions of the SIR model have been developed~\citep{bartlett1956deterministic,bailey1975mathematical,beretta1998stability,andersson2012stochastic,britton2010stochastic,kiss2017mathematics,cai2015stochastic,cai2017stochastic,karako2020analysis,martcheva2015introduction,allen2017primer,britton2019stochastic,zhou2021ergodic,laaribi2023generalized,greenwood2009stochastic}. \cite{Aurell_Carmona_Dayanıklı_Lauriere_2022b} explore a graphon game of epidemic control, with one special case involving multi-population modeling that considers different age and city groups. Their model examines the effects of different policies when individuals choose their own socialization level (control) in the game setup. However, their approach is based on an individualized version of \textit{multi-group SIR} whereas our framework focuses on the group level. In \cite{acemoglu2021optimal}, a non-homogeneous (i.e., multi-group) version of the deterministic SIR model is considered, incorporating different age groups---young, middle-aged, and old---and deriving age-specific policies to minimize overall costs. In this paper, we extend their framework to the stochastic counterpart. In addition, while the motivation behind their multi-group approach is to derive group-specific policies, our method also considers fairness, which, to the best of our knowledge, has not been explored in existing work.
 
\paragraph{Fairness in AI for Social Decision Making.} With the advancement of AI, automated decision-making and policymaking have become increasingly prevalent in critical social domains such as college admissions, loan approvals, and criminal justice. However, these AI-driven systems have raised many concerns, as they may not be entirely objective and can even exacerbate existing human biases~\citep{dastin2022amazon,angwin2022machine,samorani2022overbooked}. 
Many studies have explored fairness considerations in AI-driven decision-making across various applications. For example, \cite{azizi2018designing} study the problem of housing allocations for homeless youth and propose fair, efficient, and interpretable policies. In the context of micro-lending, \cite{liu2019personalized}~develop a fairness-aware re-ranking algorithm that balances recommendation accuracy with borrower-side fairness while also accounting for lenders’ preferences for diversity. Similarly, \cite{berk2021fairness}~provide an integrated examination of fairness and accuracy trade-offs in criminal justice risk assessments, demonstrating the inherent challenges in satisfying multiple fairness criteria simultaneously. Furthermore, \cite{kallus2022assessing}~assess disparities in lending and healthcare applications when the protected class membership is not observed in the data. They provide exact characterizations of the tightest possible set of all true disparities that are consistent with the available data. Beyond these studies, many other works examine the broader challenges and trade-offs involved in designing fair AI systems across various domains~\citep{mouzannar2019fair,corbett2023measure, kleinberg2017inherent,dai2025balancing,aghaei2019learning,raghavan2020mitigating,chen2023algorithmic,jia2024learning,wang2024wasserstein,liu2018delayed, baker2022algorithmic,taskesen2020distributionally,bertsimas2013fairness,nguyen2021scarce,corbett2017algorithmic,rahmattalabi2022learning,mashiat2022trade,rahmattalabi2021fair,freeman2020best,athanassoglou2011house}. 

\paragraph{Path Integral Control.}
Path integral control has emerged as a promising solution scheme for solving a certain class of nonlinear stochastic optimal control problems~\citep{kappen2005path}. It has recently been applied in many reinforcement learning domains, including autonomous driving~\citep{mohamed2022autonomous,williams2016aggressive,gandhi2021robust,ha2019topology}, robotics~\citep{theodorou2010generalized,chebotar2017path,williams2017information,yin2023risk,park2024distributionally,patil2022chance}, visual serving techniques~\citep{mohamed2021sampling,costanzo2023modeling,mohamed2021mppi} and finance~\citep{ingber2000high,decamps2006path,perkowski2016pathwise}. The key idea behind path integral control is to convert the value function into an expectation over uncontrolled trajectory costs. Therefore, it does not involve any optimization processes that can be intractable for solving nonlinear stochastic control problems. Instead, the method generates independent trajectories through Monte Carlo sampling and computes their associated expected costs. Furthermore, since the trajectories are independent, various parallelization techniques can be applied to significantly speed up the computation process~\citep{williams2017model}.

\paragraph{Notations.} 
Bold lower-case letter $\bm{x}\in\RR^n$ and upper-case letter $\boldsymbol{X}\in\RR^{n\times m}$ represent an $n$-dimensional vector and an $n\times m$ matrix, respectively.
$\bm{X}\in\mathbb{S}^n_{+}$ denotes an $n\times n$ positive semidefinite matrix. We define diag$(\bm{x})$ as a diagonal matrix with the vector $\bm{x}$ on its main diagonal. Similarly, diag$(\bm{X}_1,\ldots,\bm{X}_J)$ denotes a block diagonal matrix of matrices $\bm{X}_1,\ldots,\bm{X}_J$. For any $K \in \mathbb N$, we define $[K]$ as the index set $\{1,\dots,K\}$.

\section{Problem Statement}
In this section, we formalize our infectious disease mitigation problem.
\subsection{Deterministic SIR Model}  
We consider the multi-region SIR model over a continuous and finite time horizon $t \in [0, T]$. The population is partitioned into $J$ groups, where each group $j\in[J]$ represents a specific geographical region. {\color{black} Each region is characterized by different socioeconomic statuses, primarily income levels, though other heterogeneity factors may also be considered.} For each group $j\in[J]$, the spread of the infectious disease is governed by the following system of differential equations:
\begin{equation}\label{eq:dynamic}
    \begin{aligned}
    d S_j(t) & = -S_j(t) \sum_{k \in[J]} \beta_{j k} I_k(t) d t-S_j(t) V_j(t) d t \\
    d I_j(t) & =  S_j(t) \sum_{k \in[J]} \beta_{j k} I_k(t) d t-\left(\gamma_j+\delta_j\right) I_j(t) d t  -I_j(t) L_j(t) d t \\
    d R_j(t) & =\gamma_j I_j(t) d t+S_j(t) V_j(t) d t+I_j(t) L_j(t) d t \\
    d D_j(t) & =\delta_j I_j(t) d t.
    \end{aligned}
\end{equation}
Here, the state variables $\{S_j(t), I_j(t), R_j(t), D_j(t)\}$ represent the number of susceptible, infected, recovered (and immediately immune), and deceased individuals in region $j$ at time $t$. The infection rate between different regions $j,k \in [J]$ is denoted by $\beta_{j k}\left(=\beta_{k j}\right) \geq 0$ \textcolor{black}{with $\beta_{jj}$ (or simply denoted as $\beta_{j}$) representing the infection rate within region $j$}. We use $\gamma_j$ and $\delta_j$ to represent the recovery and disease-induced mortality rates of region $j$, respectively. 

In each region, policymakers implement two different control inputs to mitigate the spread of the infectious disease: vaccination and lockdown. The vaccination rate, denoted by $V_j(t)$, controls the proportion of the susceptible population $S_j(t)$ in region $j$ that acquires immunity after time $t$. The lockdown intensity, represented by $L_j(t)$, regulates the extent to which the movement and interactions of infected individuals $I_j(t)$ are restricted, thereby reducing the spread of the disease. Without loss of generality, we normalize the population in each region to be 1, i.e.,
\begin{equation*}
    S_j(t) + I_j(t) + R_j(t) + D_j(t) = 1 \quad \forall t \in [0,T]\;\;\forall j \in [J].
\end{equation*}
Adopting the standard notation in the control literature, we define the \textit{state} vector $\bm x(t) \in \mathbb{R}^{4J}$ as
\begin{equation*} 
\bm{x}(t)=[
\bm{x}_1(t),\cdots,\bm{x}_J(t)
]^\top , 
\end{equation*}
with $\bm{x}_j(t)=[
S_j(t), I_j(t), R_j(t), D_j(t)]^\top \in \mathbb{R}^4~ \;\forall j\in[J].$ 
Similarly, we define the \textit{control} vector $\bm u(t) \in \mathbb{R}^{2 J}$ as
\begin{align*}
    \bm{u}(t)=[
\bm{u}_1(t), \cdots, \bm{u}_J(t) ]^\top,
\end{align*}
with
$\bm{u}_j(t)=[
V_j(t), L_j(t) ]^\top \in \mathbb{R}^2~\forall j\in[J]$. 
Accordingly, we further define the passive (i.e., uncontrolled) {\color{black}dynamics} function $f:\mathbb{R}^{4J}\rightarrow\mathbb{R}^{4J}$ as
\begin{align*} 
f(\bm{x}(t))= [
f_1(\bm{x}_1(t)),\cdots,f_J(\bm{x}_J(t)) ]^\top , 
\end{align*}
with
\begin{equation*}
 f_j(\bm{x}_j(t)) =  \begin{bmatrix}
 -S_j(t) \sum_{k \in[J]} \beta_{j k} I_k(t)  \\
 S_j(t) \sum_{k \in[J]} \beta_{j k} I_k(t) -\left(\gamma_j+\delta_j\right) I_j(t)  \\
\gamma_j I_j(t)  \\
\delta_j I_j(t) 
\end{bmatrix} \in \mathbb{R}^4.
\end{equation*}
The control transition matrix $\bm G(\bm x(t))\in\RR^{4J\times 2J}$ is defined as $\bm G(\boldsymbol{x}(t)) = \diag(\bm{G}_1 (\bm x_1(t)), \dots, \bm{G}_J (\bm x_J(t)))$
where
\begin{align*}
    \bm G_j(\boldsymbol{x}_j(t)))=\begin{bmatrix}
-S_j(t) & 0 & S_j(t) & 0  \\
0 & -I_j(t) & I_j(t) & 0  
\end{bmatrix}^{\top} \in \mathbb{R}^{4 \times 2}.
\end{align*}
Equipped with these notations, we can now rewrite the system dynamics~\eqref{eq:dynamic} {\color{black}as follows:}
\begin{equation}\label{eq:control_dynamic}
d \boldsymbol{x}(t)=f(\boldsymbol{x}(t)) d t+\boldsymbol{G}(\boldsymbol{x}(t)) \boldsymbol{u}(\boldsymbol{x}(t)) d t.
\end{equation}
This system dynamics belong to a class of control-affine systems, which are nonlinear in the state variables due to the product term $S_j(t) I_j(t)$, but remain affine in the control inputs. The policymakers aim to determine the optimal control policy $\bm{u}^\star(t)$ for $t \in [0, T]$ by solving the following optimization problem:
\begin{equation}\label{eq:true_cost}
\min_{\bm u(\cdot)}  
\int_{0}^T
\mc{L}_{\bm u}\left(\bm x(t)\right) dt
+
\psi\left(\bm x(T)\right)
 \quad \text{s.t.}~\eqref{eq:control_dynamic}~\text{holds}.
\end{equation}
Here, the immediate cost function $\mc{L}_{\bm u}(\cdot)$ and the terminal cost function ${\psi}(\cdot)$ are defined as follows:
\begin{subequations}
\label{eq:inter_plus_terminal}
\begin{align}
&\mathcal{L}_{\bm u}\left(\bm x(t)\right)
 =
\underbrace{ q(\bm x(t)) }_{\text {state-dependent cost}}  +\underbrace{\frac{1}{2} \boldsymbol{u}(t)^{\top} \boldsymbol{R} \boldsymbol{u}(t)}_{\text {control-dependent cost }}, \label{eq:immediate_cost} \\ 
&\psi(\bm x(T)) = q(\bm x(T)).\label{eq:terminal_cost} 
\end{align}
\end{subequations}
The function $q(\cdot)$ in \eqref{eq:inter_plus_terminal} represents the state-dependent cost and is defined as
\begin{equation}
\label{eq:qt}
    q(\bm x(t)) = \underbrace{\sum_{j \in [J]} w_j\left(I_j(t)+D_j(t)\right)}_{\text {economic loss by unemployment}} + \underbrace{\eta \mathds{U}(\bm x(t)) }_{\text {unfairness penalty }}.
\end{equation}
Here, the parameter $w_j$ in \eqref{eq:qt} represents the \textcolor{black}{\textit{average pre-epidemic level of economic output}} in region $j$, and for simplicity, we assume that the unemployment rate in region $j$ is given by $I_j(t) + D_j(t)$. Consequently, the first term in $q(\cdot)$ captures the economic loss due to unemployment---a similar but more sophisticated definition of economic loss can be found in \cite{acemoglu2021optimal}, which includes several additional parameters. In addition, the control-dependent cost in \eqref{eq:immediate_cost} is a quadratic function of $\bm u(\cdot)$ and $\bm R\in\mathbb{S}^{2J}_+$ represents the control cost matrix, \textcolor{black}{which accounts for the economic and social costs associated with implementing control policies. Specifically, the cost of control reflects both the direct economic impact---such as decreased economic activity due to stricter lockdown policies or the logistical costs of vaccination distribution---as well as social costs, including the cost of life satisfaction of individuals due to decreased social interactions under stricter lockdown policies.} 

The second term $\mathds{U}(\bm x (t))$ in \eqref{eq:qt} can be any arbitrary function of the state $\bm x(t)$, serving as a measure of unfairness. Note that since the evolution of the state depends on the control $\bm u(t)$ through the dynamics in~\eqref{eq:control_dynamic}, a larger value of $\mathds U(\cdot)$ indicates a higher degree of unfairness in the resulting policy. The parameter $\eta \geq 0$ is a tunable parameter that adjusts the importance of fairness in the overall cost. In this paper, we adopt the following definition for $\mathds{U}(\cdot)$.

\begin{definition}[Economic Disparity Unfairness Measure]\label{def:unfair_measure_mu}
Given state $\bm x(t)$ at time $t$, the unfairness measure $\mathds{U}(\bm x(t))$ is defined as:
\begin{equation*}
\mathds{U}(\bm x(t)) = \displaystyle \max _{j_1, j_2 \in[J]}\left\{\left(I_{j_1}(t)+D_{j_1}(t)\right)
- \left(I_{j_2}(t)+D_{j_2}(t)\right)
\right\}.
\end{equation*}
\end{definition}
\noindent The unfairness measure in Definition \ref{def:unfair_measure_mu} is designed to reduce \textit{economic disparity} by ensuring that the maximum difference in unemployment rates between any two regions remains small. In other words, by incorporating this unfairness penalty into the cost function~\eqref{eq:qt}, we aim to derive a disease mitigation policy that prevents disproportionate unemployment rates across different socioeconomic groups.

The SIR model in~\eqref{eq:control_dynamic} is deterministic. Consequently, the sequential decision-making problem~\eqref{eq:true_cost} assumes that the control inputs lead to certain outcomes. For example, vaccinated individuals $S_j(t) V_j(t)$ become certainly immunized, or a lockdown policy perfectly isolates the fraction of the infected $I_j(t) L_j(t)$ from the susceptible.

\begin{remark}[Measures of Health Inequalities]
In the literature, there are many measures for evaluating socioeconomic inequality in healthcare. Broadly, these measures can be classified into two categories: \textit{individual level} and \textit{regional (group) level}~\citep{regidor2004measures}. Both perspectives have their advantages and limitations. In this paper, we concentrate on the \textit{group level} measures due to their straightforward definition and ease of comprehensibility for policymakers. 

We now provide a brief review of several commonly used group-level measures. \textcolor{black}{\textbf{Pairwise Comparisons}~\citep{braveman2010socioeconomic,world2013handbook} assess disparities between different groups, such as comparing the most and least wealthy populations. Historically, this has been the predominant approach in inequality monitoring due to its intuitive nature and ease of interpretation.} \textbf{Concentration Index}~\citep{wagstaff1991measurement} is calculated by comparing the cumulative percentage of the population (ranked by socioeconomic factors) against the distribution of healthcare resources. It quantifies the extent to which a health indicator is concentrated among advantaged or disadvantaged groups. \textbf{Theil Index}~\citep{theil1972statistical} evaluates the equality of health resource allocation by population across different regions. It is particularly useful for assessing relative inequalities when there is no natural ordering among different groups. Additional measures include the \textbf{Index of Dissimilarity}~\citep{pappas1993increasing} and the \textbf{Atkinson Index}~\citep{atkinson1970measurement}, both of which provide alternative perspectives on health inequality.

In practice, selecting the appropriate unfairness measure requires policymakers to consider the objectives, ethical implications, and social context. For example, \textbf{Pairwise Comparisons} could be used to assess absolute or relative disparities, which are particularly important for resource allocation. Due to space limitations, this paper focuses solely on the unfairness measure defined in Definition \ref{def:unfair_measure_mu}, which is based on Pairwise Comparisons. Nonetheless, our framework is flexible to accommodate other unfairness measures.
\end{remark}

\subsection{Stochastic SIR Model}  
The certainty of the effectiveness of control measures is a rather strong assumption. For example, the efficacy of the vaccine can have randomness due to both the vaccine formulation and the vaccination decisions of the individuals. Furthermore, even if the lockdown policies are set by a regulator, they may not be strictly followed by the individuals which in turn will create randomness in the dynamics. For this reason, we propose a stochastic counterpart of \eqref{eq:dynamic} as follows: 
\begin{equation}\label{eq:sto_dynamic}
    \begin{aligned}
    d S_j(t) & = -S_j(t) \sum_{k \in[J]} \beta_{j k} I_k(t) d t-S_j(t) \left( V_j(t) d t + d\xi_{{V_j}}(t)\right) \\
    d I_j(t) & =  S_j(t) \sum_{k \in[J]} \beta_{j k} I_k(t) d t-\left(\gamma_j+\delta_j\right) I_j(t) d t  -I_j(t)\left(L_j(t) d t + d\xi_{{L_j}}(t) \right) \\
    d R_j(t) & =\gamma_j I_j(t) d t+ S_j(t) \left( V_j(t) d t + d\xi_{{V_j}}(t)\right) + I_j(t)\left(L_j(t) d t + d\xi_{{L_j}}(t) \right) \\
    d D_j(t) & =\delta_j I_j(t) d t.
    \end{aligned}
\end{equation}
Here, $\xi_{{V_j}}(t)$ and $\xi_{{L_j}}(t)$ denote independent zero-mean Gaussian disturbances with variances $\sigma_{{V_j}}^2$ and $\sigma_{{L_j}}^2$, respectively.
Adopting the standard notation in the control literature, we define the \textit{noise} vector $\bm \xi(t) \in \mathbb{R}^{2 J}$ as 
\begin{align*}
    \bm{\xi}(t)=[
\bm{\xi}_1(t), \cdots, \bm{\xi}_J(t) ]^\top,
\end{align*}
with $\bm \xi_j(t) = [ \xi_{{V_j}}\!(t), \xi_{{L_j}}\!(t) ]^\top \in\mathbb{R}^2$ and its covariance matrix $\bm{\Sigma} \in \mathbb{S}_{+}^{2 J}$ as
$$\bm{\Sigma}= \text{diag}(\bm{\Sigma}_1, \dots, \bm{\Sigma}_J),$$
where $\bm \Sigma_j= \text{diag}(\sigma_{{V_j}}^2, \sigma_{{L_j}}^2) \in \mathbb{S}_{+}^2$. Then, similar to \eqref{eq:control_dynamic}, we can rewrite the stochastic dynamics in~\eqref{eq:sto_dynamic} {\color{black}as follows:}
\begin{equation}\label{eq:sto_control_dynamic}
d \boldsymbol{x}(t)=f(\boldsymbol{x}(t)) d t+\boldsymbol{G}(\boldsymbol{x}(t))\Big(\boldsymbol{u}(\boldsymbol{x}(t)) d t+ d \boldsymbol{\xi}(t)\Big).
\end{equation}
For $t\in[0,T]$, we define the value function
\begin{equation}\label{eq:sto_true_cost}
\mc V(\bm x(t)) = \min_{\bm u(\cdot)}  
\mathbb{E} \left[
\int_{t}^T
\mc{L}_{\bm u}\left(\bm x(s)\right)  ds
+ \psi\left(\bm x(T)\right)
\right],
\end{equation}
where $\mc{L}_{\bm u}(\cdot)$ and ${\psi}(\cdot)$ are defined the same as in~\eqref{eq:inter_plus_terminal}, and the expectation is taken over all trajectories starting at $\bm x(t)$.
Thus, policymakers seek to find the optimal policy $\bm u^\star(\bm x(t))$ for $t\in [0,T]$ under the stochastic dynamics given in~\eqref{eq:sto_control_dynamic}. 

\section{Solution Scheme}
The optimization problem~\eqref{eq:sto_true_cost} belongs to the class of nonlinear stochastic optimal control problems due to the nonlinearity of both the unfairness penalty term (Definition~\ref{def:unfair_measure_mu}) and the system dynamics~\eqref{eq:sto_control_dynamic}.
Solving nonlinear stochastic control problems, which involves solving PDEs, is generally challenging since PDEs often cannot be solved analytically, requiring numerical techniques to compute their solutions. Classical solution schemes such as finite differences, are grid-based methods that discretize the state space to obtain an approximate solution. However, the memory and computational requirements grow exponentially with the dimensionality of the state space---a phenomenon commonly known as the \textit{curse of dimensionality}. This approach becomes practically intractable when the state space dimension exceeds 3, which presents challenges given that the dimensionality of our dynamics in \eqref{eq:sto_dynamic} is even higher. To overcome the challenges posed by high dimensionality as the number of states increases, deep learning methods are also employed to solve stochastic optimal control problems and their extensions to large populations, as seen in~\cite{Gobet_Munos_2005,Han_E_2016,al2018solving,Fouque_Zhang_2020,Carmona_Lauriere_2022,Dayanikli_Lauriere_Zhang_2024}. However, these methods generally require high computational power.

Fortunately, our problem \eqref{eq:sto_true_cost} belongs to a special class of nonlinear stochastic control problems where the dynamics~\eqref{eq:sto_control_dynamic} follow a control-affine system, and the control-dependent cost is quadratic, as in~\eqref{eq:immediate_cost}. For this class of problems, a computationally efficient alternative known as \textit{path integral control} exists.
In the following, we briefly review path integral control. For further details and derivations, we refer readers to \cite{kappen2007introduction,theodorou2010generalized}.

\subsection{Path Integral Control}
 Solving the problem~\eqref{eq:sto_true_cost} involves
 setting up the following second-order PDE, known as the stochastic Hamilton-Jacobi
Bellman (HJB) equation~\citep{stengel1994optimal,fleming2006controlled}:
\begin{equation}
\label{eq:HJB}
\begin{aligned}
     -{\partial_t \mc V(\bm x(t))} 
        &=
        \min _{\bm u(\cdot)} 
     \left( q + \frac{1}{2}\bm u^{\top} \bm R \bm u
     + \partial_{\bm x}\mc V^{\top} (f + \bm G \bm u )  + \frac{1}{2}\operatorname{tr} \left( 
     \partial_{\bm x\bm x} \mc V (\bm G \bm \Sigma \bm G ^\top)   
    \right) 
    \right),
\end{aligned}
\end{equation}
with boundary condition $\mc V(\bm x(T))=\psi(\bm x(T))$. 
Here, \(\partial_t \mathcal{V}\) and \(\partial_{\bm{x}} \mathcal{V}\) denote the partial derivatives of the value function \(\mathcal{V}(\bm x(t))\) with respect to time \(t\) and the state vector \(\bm{x}(t)\), respectively, while \(\partial_{\bm{x}\bm{x}} \mathcal{V}\) represents the second-order partial derivative with respect to $\bm x(t)$. For notational simplicity, we sometimes suppress the dependence of functions on $\bm x(t)$ on the right-hand side, e.g., \(q = q(\bm{x}(t))\), \(\partial_{\bm{x}} \mathcal{V} = \partial_{\bm{x}} \mathcal{V}(\bm{x}(t))\), \(\bm G=\bm G(\bm x(t))\), and similarly for other terms.  
Note that the minimization in \eqref{eq:HJB} is a convex quadratic optimization problem. Taking the derivative with respect to $\bm u(t)$ on the right-hand side in \eqref{eq:HJB} and setting it to zero, one can find the corresponding optimal control:
\begin{equation}\label{eq:optimal_control}
\bm u^\star (\bm x(t)) = - \bm R^{-1} \bm G(\bm x(t))^\top {\partial_{\bm x} \mc V(\bm x(t))}. 
\end{equation}
Substituting \eqref{eq:optimal_control} into \eqref{eq:HJB}, we obtain:
\begin{equation}
\label{eq:HJB2}
\begin{aligned}
    -{\partial_t \mc V} 
   &  = q  +
     \partial_{\bm x}\mc V^{\top}f 
     - 
     \frac{1}{2} \partial_{\bm x}\mc V^\top (\bm G \bm R^{-1} \bm G^\top) \partial_{\bm x}\mc V + \frac{1}{2}\operatorname{tr} \left( 
     \partial_{\bm x\bm x} \mc V (\bm G \bm \Sigma \bm G ^\top )  
    \right).
\end{aligned}
\end{equation}
In order to find a solution to the PDE above, we use a logarithmic transformation of the value function:
\begin{equation}\label{eq:log_transform}
    \mathcal{V}(\bm{x}(t)) = -\lambda \log \phi(\bm{x}(t)).
\end{equation}
Given this logarithmic transformation, the HJB equation \eqref{eq:HJB2} yields the following:
\begin{equation}
\label{eq:HJB3}
\begin{aligned}
    \frac{\lambda}{\phi}\partial_t \phi 
    & =
     q - \frac{\lambda}{\phi} \partial_{\bm x}\phi^{\top} f 
     - \frac{\lambda^2}{2\phi^2} \partial_{\bm x}\phi^{\top} \bm G \bm R^{-1} \bm G ^\top \partial_{\bm x}\phi + \frac{\lambda}{2\phi^2} \operatorname{tr}(\partial_{\bm x}\phi^{\top} \bm G \bm \Sigma \bm G \partial_{\bm x}\phi)  - \frac{\lambda}{2\phi} \operatorname{tr}(\partial_{\bm x \bm x}\phi (\bm G \bm \Sigma \bm G^{\top})).
\end{aligned}
\end{equation}
Conventional approaches to solving the HJB equation~\eqref{eq:HJB3} involve backward evaluation of the value functions over the entire time horizon $[0, T]$ for all $\bm x(t)$. This process requires numerically discretizing the continuous state space into a grid, where the level of precision determines the number of points on the grid. As mentioned earlier, this recursive backward evaluation suffers from the curse of dimensionality, i.e., the number of grid points grows exponentially with the dimension of the state space, making the approach computationally intractable for high-dimensional problems.

Path integral control can be an alternative to the backward recursion for solving problem~\eqref{eq:HJB3}. We assume that the control cost matrix satisfies the following condition:
\begin{equation}\label{eq:condition}
    \exists \lambda\geq 0, \text{ s.t } \lambda \bm R^{-1} =  \bm \Sigma.
\end{equation}
With \eqref{eq:condition}, the HJB equation~\eqref{eq:HJB3} simplifies to the following form:
\begin{equation}
\label{eq:HJB4}
        -{\partial_t \phi} 
        =
     -\frac{1}{\lambda}q\phi 
     +
     \partial_{\bm x}\phi ^\top f
     + \frac{1}{2}\operatorname{tr} \left( 
     \partial_{\bm x\bm x} \phi (\bm G \bm \Sigma \bm G ^\top )  
    \right),
\end{equation}
with boundary condition $\phi(\bm x(T))=\exp\left(-\frac{1}{\lambda}\psi(\bm x(T))\right)$. Note that the transformed HJB equation~\eqref{eq:HJB4}, known as the backward Chapman-Kolmogorov PDE, is linear in $\phi(\cdot)$. Subsequently, the linearity allows for applying the Feynman-Kac {\color{black}theorem}. (Theorem~8.2.1 in \cite{oksendal2013stochastic}), yielding the solution to~\eqref{eq:HJB4} as follows:
\begin{equation}
\label{eq:HJB_solution}
\begin{aligned}
    \phi(\bm x(t)) 
    = 
    \mathbb{E}\left[
        \exp
        \left(
        -\frac{1}{\lambda}\mc J_t(\overline{\bm x})
        \right)
    \right],
\end{aligned}
\end{equation}
where $\overline{\bm x}$ represents \emph{uncontrolled} dynamics which observes ~\eqref{eq:sto_control_dynamic} starting from ${\bm x}(t)$ at time $t$ with $\bm u(s)=\bm 0$ for $s\in[t,T]$, and $\mc J_t(\overline{\bm x}) = \int^T_t q(\overline{\bm x}(s))ds + \psi(\overline{\bm x}(T))$ represents the associated trajectory cost. Notably, by applying the Feynman-Kac {\color{black}theorem}, evaluating the value function---i.e., solving the HJB equation~\eqref{eq:HJB2}---amounts to computing the expectation on the right-hand side of~\eqref{eq:HJB_solution} and then transforming $\phi(\cdot)$ back to $\mc V(\cdot)$ via~\eqref{eq:log_transform}. This underpins the key idea of path integral control: transforming the problem of solving the nonlinear PDE into computing the expectation of trajectory costs.

As further established in~\cite{theodorou2015nonlinear}, taking the derivative of $\phi(\cdot)$ with respect to $\bm{x}(t)$ yields the following optimal control, given the current state $\bm{x}(t)$ at time $t \in [0,T]$
\begin{equation}\label{eq:optimal_control2}
\bm u^\star (\bm x(t)) =
\mc G(\bm x(t))
\frac{
\displaystyle
\mathbb{E}\left[
        \exp
        \left(
        -\frac{1}{\lambda}\mc J_t(\overline{\bm x})
        \right)\bm G_c(\bm x(t))d\bm\xi(t)
    \right] }
{\displaystyle\mathbb{E}\left[
        \exp
        \left(
        -\frac{1}{\lambda}
        \mc J_t(\overline{\bm x})
        \right)
    \right]},
\end{equation}
where 
\begin{equation}\label{eq:mc_G_matrix}
    \mc G(\bm x(t))=\bm R^{-1} \bm G_c^\top(\bm x(t))(\bm G_c(\bm x(t))\bm R^{-1}\bm G_c(\bm x(t))^\top)^{-1}.
\end{equation}
Here, $\bm G_{c}(\cdot)\in\mathbb{R}^{3J\times 2J}$ denotes the submatrix of the control transition matrix $\bm G(\cdot)\in\RR^{4J\times 2J}$, corresponding to the directly actuated states, denoted by $\bm x_c(t)\in\mathbb{R}^{3J}$. The overall state vector is partitioned as
$\bm x(t)=[\bm x_c(t)^\top \; \bm x_p(t)^\top]^\top$,
where $\bm x_p(t)\in\mathbb{R}^{J}$ represents the non-directly actuated states.

\begin{remark}
    The assumption in~\eqref{eq:condition} ensures the linearity of the transformed HJB equation~\eqref{eq:HJB3}. As shown in~\cite{kappen2005path}, this assumption implies that in directions with low noise, control is expensive, with the cost approaching infinity as the variance of noise tends to zero. This interpretation is particularly relevant to our disease mitigation problem. For instance, achieving extremely high precision in vaccination effectiveness would be exceedingly costly, as it would require extensive efforts such as rigorous and frequent testing. A similar rationale applies to lockdown policies, where enforcing strict containment measures with absolute precision would demand excessive resources and logistical efforts.
\end{remark}

\subsection{Numerical Method}\label{sec:numerical_method}
Note that~\eqref{eq:optimal_control2} represents the optimal control in a continuous-time, continuous-state space. To numerically implement path integral control, two types of approximations are required: time discretization and trajectory sampling. 

First, applying the Euler-Maruyama method~\citep{maruyama1955continuous}, we obtain the discrete-time version of the dynamics \eqref{eq:sto_control_dynamic} as follows:
\begin{equation}
\label{eq:time_discretized_dynamic_main}
\bm x_{k+1}= \bm x_k + f(\bm x_k)\Delta t+ \bm G(\bm x_k)\Big(\bm u(\bm x_k)\Delta t
+ 
\bm \epsilon \sqrt{\Delta t} 
\Big),
\end{equation}
for $k=0,1,\ldots,K-1$. Here, the step size $\Delta t>0$ determines the number of time steps, resulting in $K+1$ steps where $K=T/\Delta t$, and $\bm \epsilon\sim \mathcal{N}(\bm 0, \bm \Sigma)$ is the discrete-time Gaussian noise. 
Accordingly, the trajectory cost $\mc J_t(\overline{\bm x})$ in \eqref{eq:optimal_control2} is approximated as follows:
for $t=0, \Delta t, \ldots, (K-1)\Delta t$,
\begin{equation*}
\mc J_t(\overline{\bm x}) \approx \displaystyle \sum_{k=t/\Delta t}^{K-1} q(\overline{\bm x}_k)\Delta t + \psi(\overline{\bm x}_{K}). 
\end{equation*}
Since computing the expectation $\mathbb{E}[\mc J_t(\overline{\bm x})]$ in \eqref{eq:optimal_control2} requires approximation, we employ the Monte Carlo method. Specifically, we generate $M$ sample trajectories $\{\overline{\bm x}_{k}^{(m)}\}_{k=0}^K$ for $m=1,\ldots,M$ based on the discretized system dynamics in~\eqref{eq:time_discretized_dynamic_main}. The expected trajectory cost is then approximated as
\begin{equation*}
\begin{aligned}
\mathbb{E}[\mc J_t(\overline{\bm x})] 
&\approx
\sum_{m=1}^M\left(
\sum_{k=t/\Delta t}^{K-1} q(\overline{\bm x}_k^{(m)}) \Delta t+ \psi(\overline{\bm x}_{K}^{(m)})  
\right)  = \sum_{m=1}^M \widehat{\mc J}^{(m)}_t,
\end{aligned}
\end{equation*}
where, for notational convenience, we define $\widehat{\mc J}^{(m)}_{t}=\sum_{k=t/\Delta t}^{K-1}q(\overline{\bm x}_k^{(m)})\Delta t + \psi(\overline{\bm x}_{K}^{(m)})$ as the $m$-th trajectory cost. Consequently, the optimal control~\eqref{eq:optimal_control2} can be approximated as follows. For $t=0, \Delta t, \ldots, (K-1)\Delta t$, we have
\begin{equation}\label{eq:approx_control_main}
\bm u^\star (\bm x(t)) 
\approx
\mc G(\bm x(t)) \frac{
\displaystyle
\sum^{M}_{m=1}
        \exp
        \left(
        -\frac{1}{\lambda}\widehat{\mc J}^{(m)}_t 
        \right)\frac{\bm G_c(\bm x(t))\bm \epsilon^{(m)}}{\sqrt{\Delta t}}
}
{
\displaystyle
\sum^{M}_{m=1}
        \exp
        \left(
        -\frac{1}{\lambda}\widehat{\mc J}^{(m)}_t 
        \right)
},
\end{equation}
where $\mc G(\bm x(t))$ is the same as~\eqref{eq:mc_G_matrix}.

\section{Numerical Case Study}
In this section, we present a case study on COVID-19 to derive managerial insights using our proposed approach. Specifically, we apply the path integral algorithm to solve the stochastic optimal control problem with the underlying stochastic multi-region SIR model~\eqref{eq:sto_dynamic}. We then compare the resulting region-specific policies with a homogeneous policy obtained from the standard single-group SIR model (i.e., $J=1$). The primary objectives of this case study are twofold:
\begin{enumerate}[label=\textbf{\roman*})]
    \item Comparison of Different Policy Frameworks: we compare the region-specific policies obtained from the multi-group SIR model with the homogeneous policy derived from the single-group SIR model.
    \item Impact of Fairness Considerations: we examine how varying levels of $\eta$ influence the optimal policy and the overall effectiveness in disease mitigation.
\end{enumerate}
All experiments were implemented in Python 3.7 and conducted on a laptop equipped with a 6-core, 2.3 GHz Intel Core i7 CPU and 16 GB of RAM.

\begin{table}[htbp]
\centering
\begin{tabular}{@{}lccccc@{}}
\toprule
 & \multicolumn{5}{c}{\textbf{Group-specific Parameters}} \\
\cmidrule(lr){2-6}
 & \multirow{2}{*}{$\beta_j$} & \multirow{2}{*}{$\gamma_j$} & \multirow{2}{*}{$\delta_j$} & \multirow{2}{*}{$w_j$} & \textbf{Initial State} \\
 \cmidrule(lr){6-6}
 & & & & & $\{S_j(0), I_j(0), R_j(0), D_j(0)\}$ \\
\midrule
Upper  & 0.2  & 0.1 & 0.03 & 2.0   & \{0.99, 0.01, 0.0, 0.0\} \\
Middle & 0.2  & 0.1 & 0.03 & 1.0   & \{0.99, 0.01, 0.0, 0.0\} \\
Lower  & 0.3  & 0.1 & 0.05 & 2/3   & \{0.99, 0.01, 0.0, 0.0\} \\ \midrule
Single & 0.23 & 0.1 & 0.03 & 1.2   & \{0.99, 0.01, 0.0, 0.0\} \\
\bottomrule
\end{tabular}
\vspace{0em}  
\begin{tabular}{@{}cccccc@{}} 
\toprule
\multicolumn{6}{c}{\textbf{Other Parameters}} \\ 
\cmidrule(lr){1-6}
    $T$             & $\Delta t$   & $\sigma_V$ & $\sigma_L$ & $M$ (\text{\# of Sample Trajectories in} \eqref{eq:approx_control_main}) &   \\
\midrule
 180~\text{days}  & 1~\text{day} & 0.01       & 0.01  & 1000  & \\
\bottomrule
\end{tabular}
\caption{Parameter values for the (single-/multi-region) SIR models and the path integral control algorithm.}
\label{tab:parameters}
\end{table}

\begin{figure*}[htb!]
    \centering
    \includegraphics[width=\textwidth]{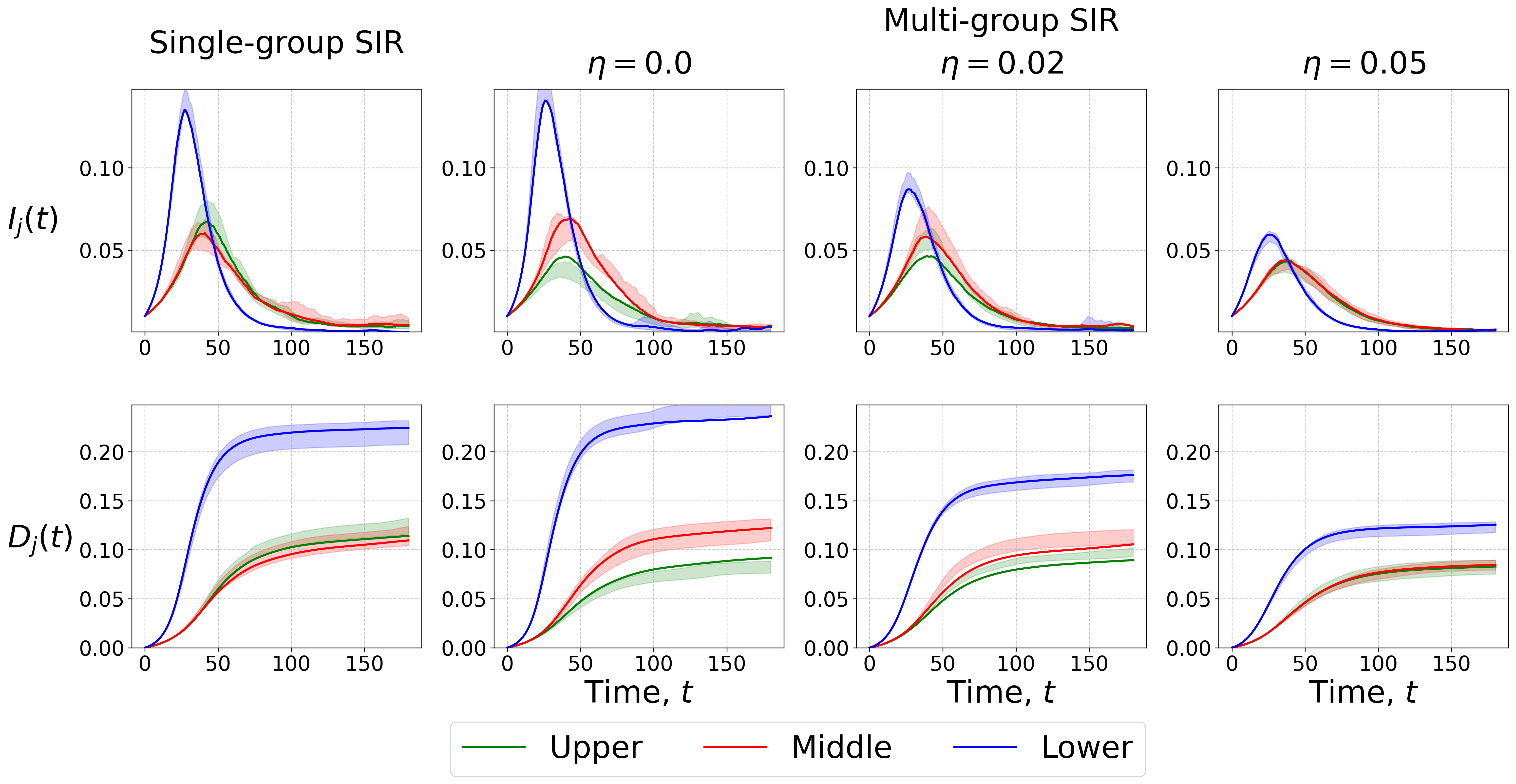} 
    \caption{
        Mean evolution (solid lines) of the infected $I_j(t)$ (first row) and deceased $D_j(t)$ (second row) over 500 simulations across different regions---upper (green), middle (red), and lower (blue) income groups with shaded areas representing the 10th and 90th percentiles: the first column presents test performance under the homogeneous policy based on the single-group SIR model, while the remaining columns show results for our region-specific policy derived from the multi-group SIR model with varying penalty parameter $\eta$. Increasing $\eta$ significantly mitigates the effects of the disease in the lower-income region.
    }
    \label{fig:state_comparison_I_D_only}
\end{figure*}

\paragraph{Experiment Setup.}
We categorize the population into three income-based groups: upper, middle, and lower. This categorization reflects the socioeconomic disparities observed during COVID-19, where lower-income groups experienced more severe financial struggles and higher mortality rates compared to wealthier groups. The parameter values for our case study, based on the stochastic multi-region SIR model~\eqref{eq:sto_dynamic}, are presented in Table~\ref{tab:parameters}. For middle- and upper-income level regions, we adopt parameter ranges commonly used in the literature. However, to reflect socioeconomic disparities, several parameters are adjusted for the lower-income region.
Due to space limitations, we relegate detailed discussions on the choice of parameter values in the appendix~\ref{supp:parameters}.

\begin{figure*}[ht!]
    \centering
    \includegraphics[width=\textwidth]{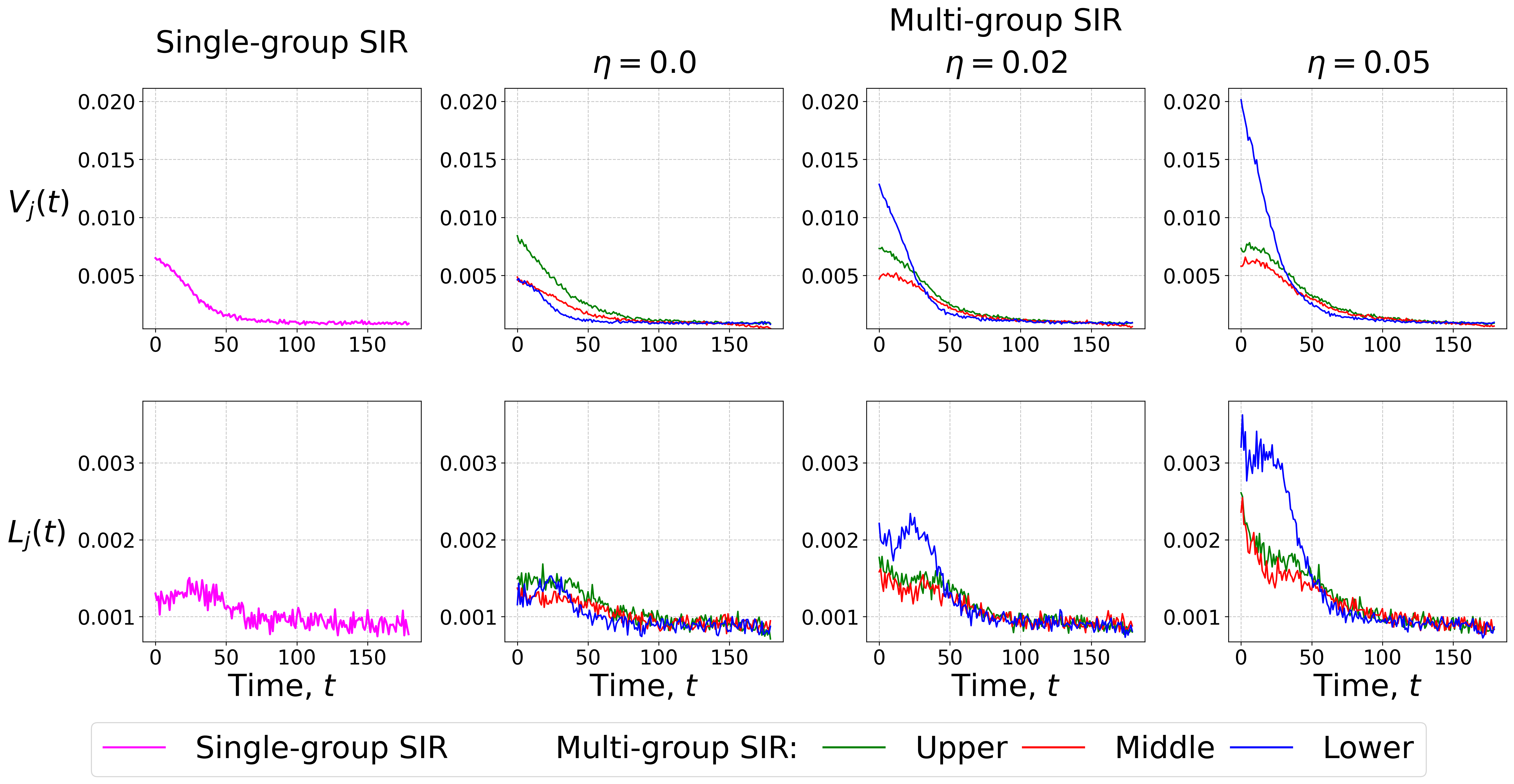} 
      \caption{
      Mean control inputs for vaccination $V_j(t)$ (first row) and lockdown $L_j(t)$ (second row) policies over 500 simulations: the first column shows the mean control input (purple) under the homogeneous policy based on the single-group SIR model. The remaining columns represent our region-specific policy for the multi-group SIR model with varying penalty parameters $\eta$ across different regions—upper (green), middle (red), and lower (blue) income groups. Increasing $\eta$ results in significantly different vaccination policies, particularly in the lower-income region.
      }
    \label{fig:control_comparison}
\end{figure*}
\noindent We consider a time horizon of $T = 180$ days with a time step of $\Delta t = 1$ day. Hence, the total number of time steps is 181, where $K = T / \Delta t = 180$, which is sufficient to observe the converging behavior of the dynamics in our experiments. At each time step $t = 1,\ldots,180$, we generate $1000$ uncontrolled trajectories---i.e., $M = 1000$ in \eqref{eq:approx_control_main}---starting from the current state $\bm x(t)$ and approximate the optimal control $\bm{u}^\star(\bm{x}(t))$ as in \eqref{eq:approx_control_main}. The computation of \eqref{eq:approx_control_main} takes only 0.03 seconds. Further discussions on computation times are provided in Appendix~\ref{supp:comp_time}. To assess the influence of fairness considerations, we conducted 500 simulations across different values of the fairness control parameter $\eta$, varying from 0 to 0.08. When $\eta = 0$, the optimal control inputs correspond to {\color{black}a policy} that disregards the unfairness measure $\mathds{U}(\cdot)$ in~\eqref{eq:inter_plus_terminal}.

\paragraph{Infection and Mortality Trends.} 
Figure~\ref{fig:state_comparison_I_D_only} presents the mean evolution of infected and deceased populations under different policies. The results indicate that while {\color{black}the policy that disregards the unfairness measure} is similar to the homogeneous policy, policies with higher fairness parameters $\eta$ substantially reduce disparities in infection and mortality rates between the low-income region and other regions. 
\paragraph{Variations in Optimal Policies.}
In Figure \ref{fig:control_comparison}, we compare the mean control inputs across different policies. A key observation is that vaccination policies exhibit significant variations with respect to $\eta$, whereas lockdown measures generally increase as $\eta$ grows. Specifically, under {\color{black}the policy that disregards the unfairness measure}, vaccination efforts are prioritized in the upper-income region, whereas fairness-aware policies shift the focus toward the low-income region. In fact, as $\eta$ increases, the vaccination in the upper-income region decreases. This result likely arises because the lower-income region experiences higher infection and mortality rates, making vaccination a more effective intervention for mitigating disease compared to lockdown policy in the region. 
\begin{figure}[htb!]
    \centering
    \includegraphics[width=.5\textwidth]{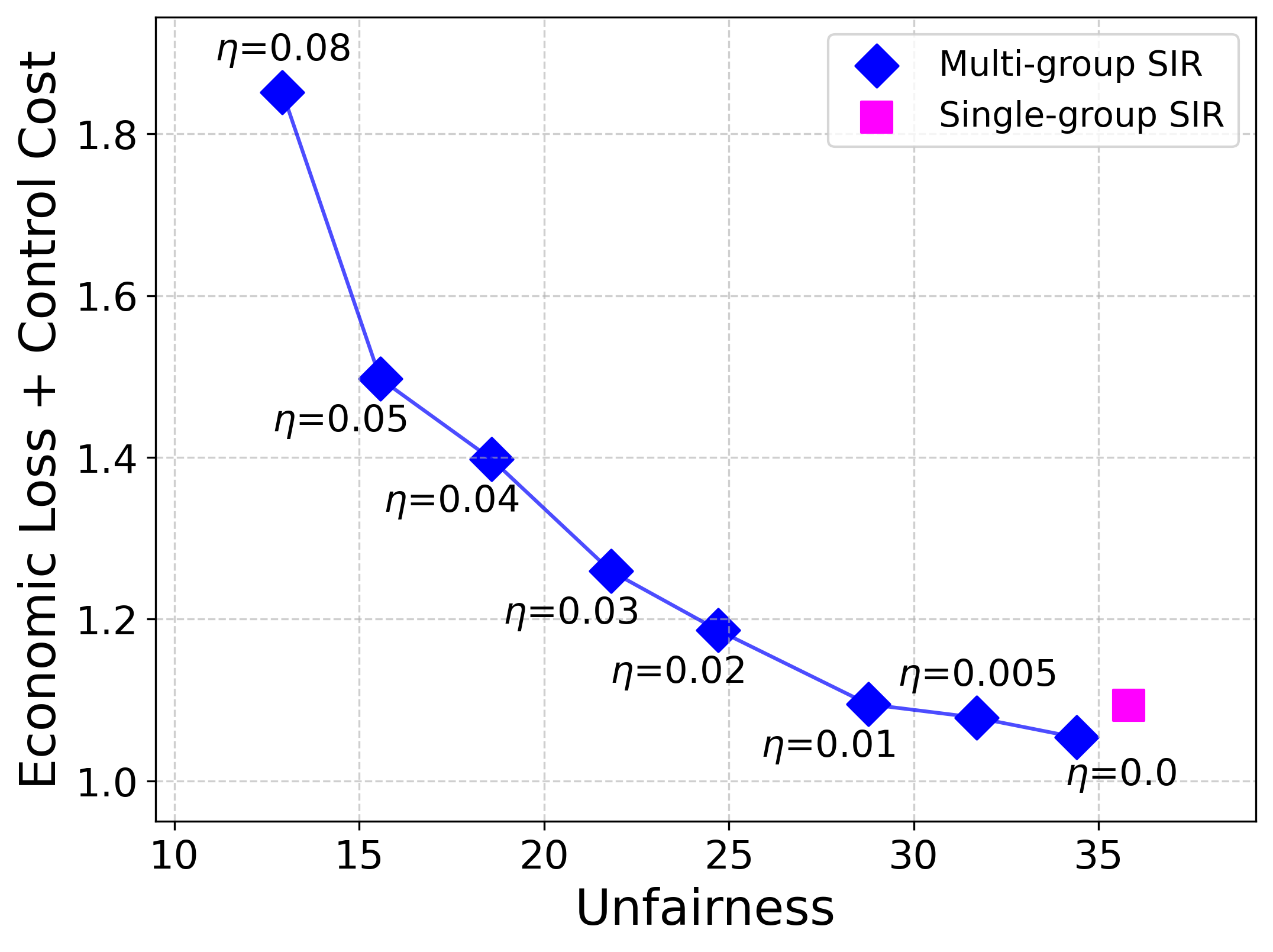} 
    \caption{
    Cost-unfairness Pareto frontier: mean values of economic loss plus control cost and unfairness measure for our multi-region SIR model (blue) over 500 simulations are shown. For example, comparing with the homogeneous policy (purple), the policy that ignores the unfairness measure ($\eta=0$) achieves better performance in both fairness and costs, i.e., a lower unfairness measure and costs. There is a clear trade-off between costs and fairness. The relatively flat curve up to $\eta=0.05$ suggests that fairness can be significantly improved with small additional costs.
    }
    \label{fig:pareto_frontier}
\end{figure}
\paragraph{Trade-off Between Fairness and Costs.} To further analyze the relationship between fairness and costs, we plot the cost-unfairness Pareto frontier in Figure~\ref{fig:pareto_frontier}. This figure illustrates the trade-off between total costs---including economic loss and control costs---and fairness across policies derived from the multi-group model. Note that, for the homogeneous policy, we display only a single result, as the single-group SIR model cannot incorporate the unfairness penalty. As expected, the homogeneous policy results in poor fairness. Interestingly, {\color{black}the policy that disregards the unfairness measure} outperforms the homogeneous policy in both costs and fairness, although the improvement is marginal.  

More importantly, fairness-aware policies with moderate values of $\eta$ achieve substantial improvements in fairness with only a marginal increase in costs. For instance, the policy with $\eta = 0.01$ achieves a $20\%$ improvement in relative fairness with virtually no additional costs. These results suggest that our region-specific policies have clear benefits, even in cases where policymakers are reluctant to sacrifice additional costs for fairness.

\paragraph{Acknowledgements} 
G\"{o}k\c{c}e Dayan{\i}kl{\i} is supported by the National Science Foundation under grant DMS-2436332. Grani A.~Hanasusanto is supported by the National Science Foundation under grants 2343869 and 2404413.

\bibliographystyle{plainnat}
\bibliography{references}

\newpage

\clearpage
\appendix
\section*{Appendices}
\section{Further Experiment Setup}
\label{supp:parameters}  
\paragraph{Parameters for Multi-group SIR Model.}
We categorize the population into three income-based groups: upper, middle, and lower. To reflect socioeconomic disparities between lower-income groups and wealthier groups, we adjusted several parameters in the stochastic multi-group SIR model~\eqref{eq:sto_dynamic} and the average pre-epidemic level of economic output $w_j$ in \eqref{eq:qt}:

\begin{enumerate}
    \item \textbf{Infection Rates}:
    The lower-income group’s infection rate is set to $\beta_{\text{lower}} = 0.3$, compared to $\beta_{\text{upper}} = \beta_{\text{middle}} = 0.2$ for wealthier groups. This reflects higher risks observed during COVID-19. As reported in ~\cite{bambra2020covid,ronkko2022impact,masterson2023disparities}, crowded living conditions, reliance on public transportation, and high-contact occupations leads to higher infection rates in the lower-income groups. For simplicity, cross-group infection rates are simplified to $\beta_{ij}=0$ (no interactions between regions).

    \item \textbf{Mortality Rates}: 
    The lower-income group’s mortality rate is increased to $\delta_{\text{lower}} = 0.05$, from $\delta_{\text{upper}} = \delta_{\text{middle}} = 0.03$ for other groups, accounting for limited access to healthcare. 

    \item \textbf{Average Pre-epidemic Level of Economic Output}: 
    The economic output $w_j$ for each group is scaled by income levels. The middle-income group's daily contribution ($w_{\text{middle}} = 1.0$) serves as a baseline. Following the income classification framework in~\cite{pew2021income}, we set the lower-income group’s contribution at two-thirds of the baseline ($w_{\text{lower}} = 2/3$), while the upper-income group contributes twice the baseline amount ($w_{\text{upper}} = 2.0$).
\end{enumerate}
\paragraph{Parameters for Single-group SIR Model.}
Most common approaches for learning disease mitigation policies are based on the classical SIR model, which we refer to as the single-group SIR model, as it is a special case of our multi-group version~\eqref{eq:sto_dynamic} when $J=1$. In this case, only a single set of parameters, $\{\beta, \delta, \gamma, w\}$, needs to be selected. We assume that the policymaker uses the average of the group-specific parameters for the single-group SIR model.  

All parameters are summarized in Table~\ref{tab:parameters}.

\newpage
\section{Computation Time}
\label{supp:comp_time}  

As discussed in Section~\ref{sec:numerical_method}, the path integral control framework approximates the optimal control input via Monte Carlo sampling. Specifically, at each time step $k=0,\ldots,K-1$, the main computational task is to generate $M$ trajectories $\{\bm{x}^{(m)}_{s}\}_{s=k}^{K}$ $\forall m\in[M]$ following the discrete-time dynamics in \eqref{eq:time_discretized_dynamic_main}, and then compute the approximated control in \eqref{eq:approx_control_main}. The accuracy of the approximation improves as the number of generated trajectories increases, hence, the trade-off between computation and performance.

\begin{figure}[htb!]
    \centering
    \includegraphics[width=0.7\textwidth]{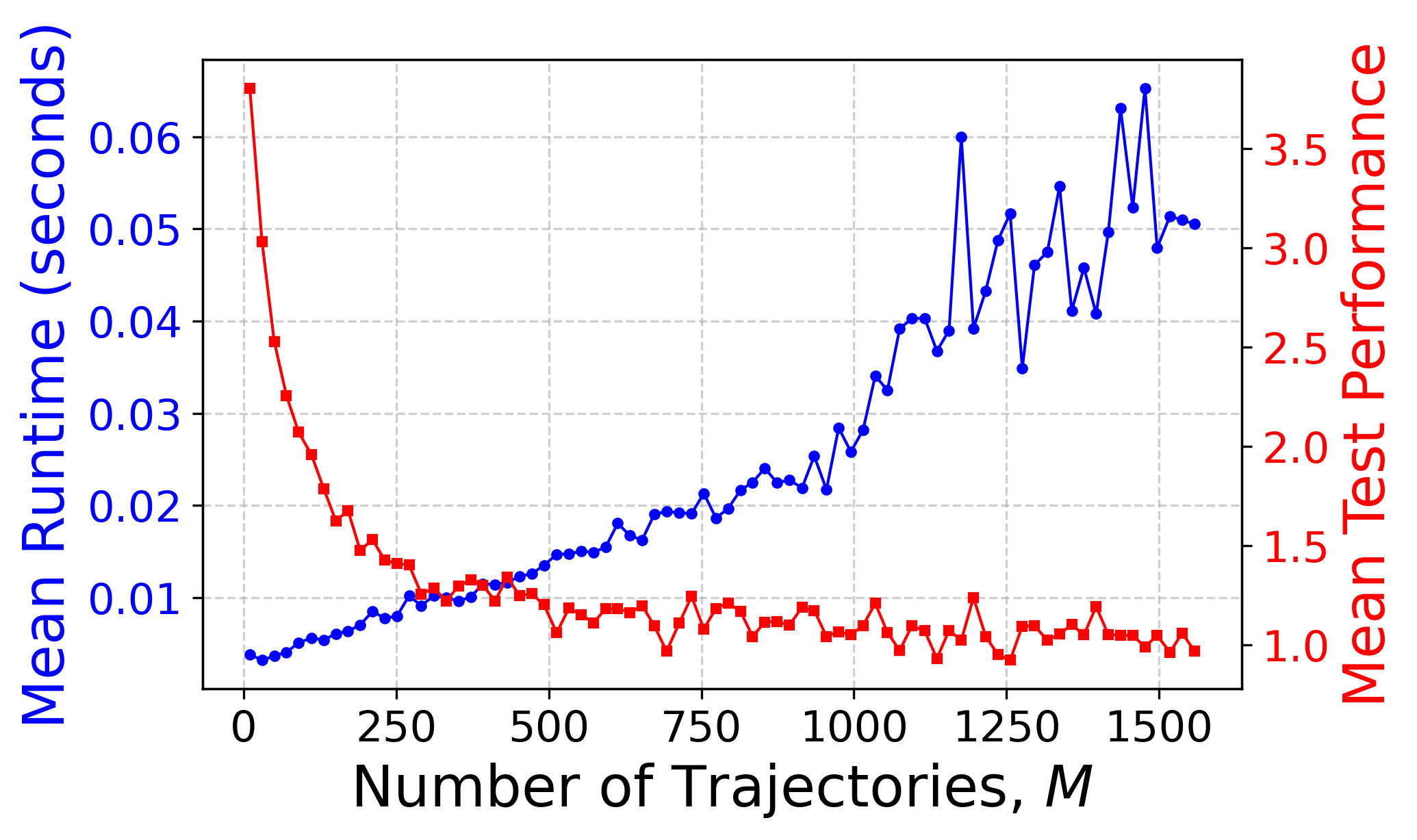} 
    \caption{
    Mean runtime (blue) for computing \eqref{eq:approx_control_main} at the initial time step ($k=0$) with $K=180$ (as in our case study) and mean test performance (red) over 30 simulations with varying numbers of sample trajectories $M$.
    }
    \label{fig:computation}
\end{figure}

Since the trajectories are generated independently, we parallelize trajectory generation on a multi-core CPU, significantly accelerating the computation process. Figure~\ref{fig:computation} reports the mean runtime for computing \eqref{eq:approx_control_main} at the initial time step—--i.e., generating $M$ sample trajectories $\{\bm x_{s}^{(m)}\}_{s=0}^{180}$ of length $181$ plus other required computations--— and the corresponding mean test performance (i.e., the mean out-of-sample objective value under the resulting control inputs) for different values of $M$. As shown, computation remains under a fraction of a second regardless of $M$, while test performance improves significantly for $M > 250$.

For robotics applications requiring real-time control, where $\Delta t$ is fractional seconds, modern GPUs can be utilized for even greater speedup~\citep{williams2017model}. However, this level of acceleration is not necessary in the context of our disease mitigation problem.

\end{document}